\documentclass[10pt,twocolumn,letterpaper]{article}

\usepackage{cvpr}              
\usepackage{makecell}
\usepackage{pifont}
\usepackage{multirow}
\usepackage{bm}
\usepackage{enumitem}
\usepackage{siunitx}
\usepackage{xcolor}
\usepackage{flushend}
\usepackage[accsupp]{axessibility}  
\definecolor{cvprblue}{rgb}{0.21,0.49,0.74}
\usepackage[pagebackref,breaklinks,colorlinks,allcolors=cvprblue]{hyperref}

\def\paperID{6976} 
\def\confName{CVPR}
\def\confYear{2025}

\title{SVLTA: Benchmarking Vision-Language Temporal \\ Alignment via Synthetic Video Situation}

\author{Hao Du$^{1}$\thanks{Both authors contributed equally.}
\and
Bo Wu$^{2}$\footnotemark[1] 
\and
Yan Lu$^{3}$ 
\and
Zhendong Mao$^{1}$\thanks{Corresponding author.} 
\and
$^1$ University of Science and Technology of China \\
$^2$ MIT-IBM Watson AI Lab \\
$^3$ The Chinese University of Hong Kong \\
\small{
\href{https://svlta-ai.github.io/SVLTA}{https://svlta-ai.github.io/SVLTA}
}\\
\small{
dh97@mail.ustc.edu.cn, \quad 
bo.wu@ibm.com, \quad
yanlu@cuhk.edu.hk, \quad
zdmao@ustc.edu.cn
} \\
}

\begin{document}
\renewcommand{\baselinestretch}{0.98}
\maketitle
\begin{abstract}
Vision-language temporal alignment is a crucial capability for human dynamic recognition and cognition in real-world scenarios. While existing research focuses on capturing vision-language relevance, it faces limitations due to biased temporal distributions, imprecise annotations, and insufficient compositionally.
To achieve fair evaluation and comprehensive exploration, our objective is to investigate and evaluate the ability of models to achieve alignment from a temporal perspective, specifically focusing on their capacity to synchronize visual scenarios with linguistic context in a temporally coherent manner. 
As a preliminary step, we present the statistical analysis of existing benchmarks and reveal the existing challenges from a decomposed perspective.
To this end, we introduce \textbf{SVLTA}, the \underline{S}ynthetic \underline{V}ision-\underline{L}anguage \underline{T}emporal \underline{A}lignment derived via a well-designed and feasible control generation method within a simulation environment. 
The approach considers commonsense knowledge, manipulable action, and constrained filtering, which generates reasonable, diverse, and balanced data distributions for diagnostic evaluations. 
Our experiments reveal diagnostic insights through the evaluations in temporal question answering, distributional shift sensitiveness, and temporal alignment adaptation. 
\end{abstract}

\section{Introduction}
\label{sec:intro}

Multimodal Large Language Models (MLLMs)~\cite{alayrac2022flamingo,li2023blip2,lin2023video} are pioneering a new direction beyond LLMs~\cite{touvron2023llama,touvron2023llama2,chiang2023vicuna}.
They have demonstrated remarkable advancements in mainstream evaluations including vision-language comprehension~\cite{liu2023visual,huang2023language,ye2023mplug,zhu2023minigpt}, analysis~\cite{alayrac2022flamingo,li2023seed,yue2023mmmu}, alignment~\cite{radford2021learning,cherti2023reproducible,su2020adapting}, and even reasoning~\cite{wu2021star,urooj2023learning}.
However, current assessments primarily focus on the models' performances of semantic-aligned vision-language inferences, often neglecting if models perform well on the capacities in real-world situations that evolve over time. This temporal perspective is crucial for effective human cognition and adaptation to the surrounding environment~\cite{buch2022revisiting,bagad2023test}.
In this work, we investigate and assess the ability of models to achieve vision-language temporal alignment, specifically how they can synchronize visual events with linguistic context in a temporal manner. This remains a significant challenge for the comprehensive evaluation of multimodal models.

\begin{figure*}[t]
    \includegraphics[width=0.9\linewidth]{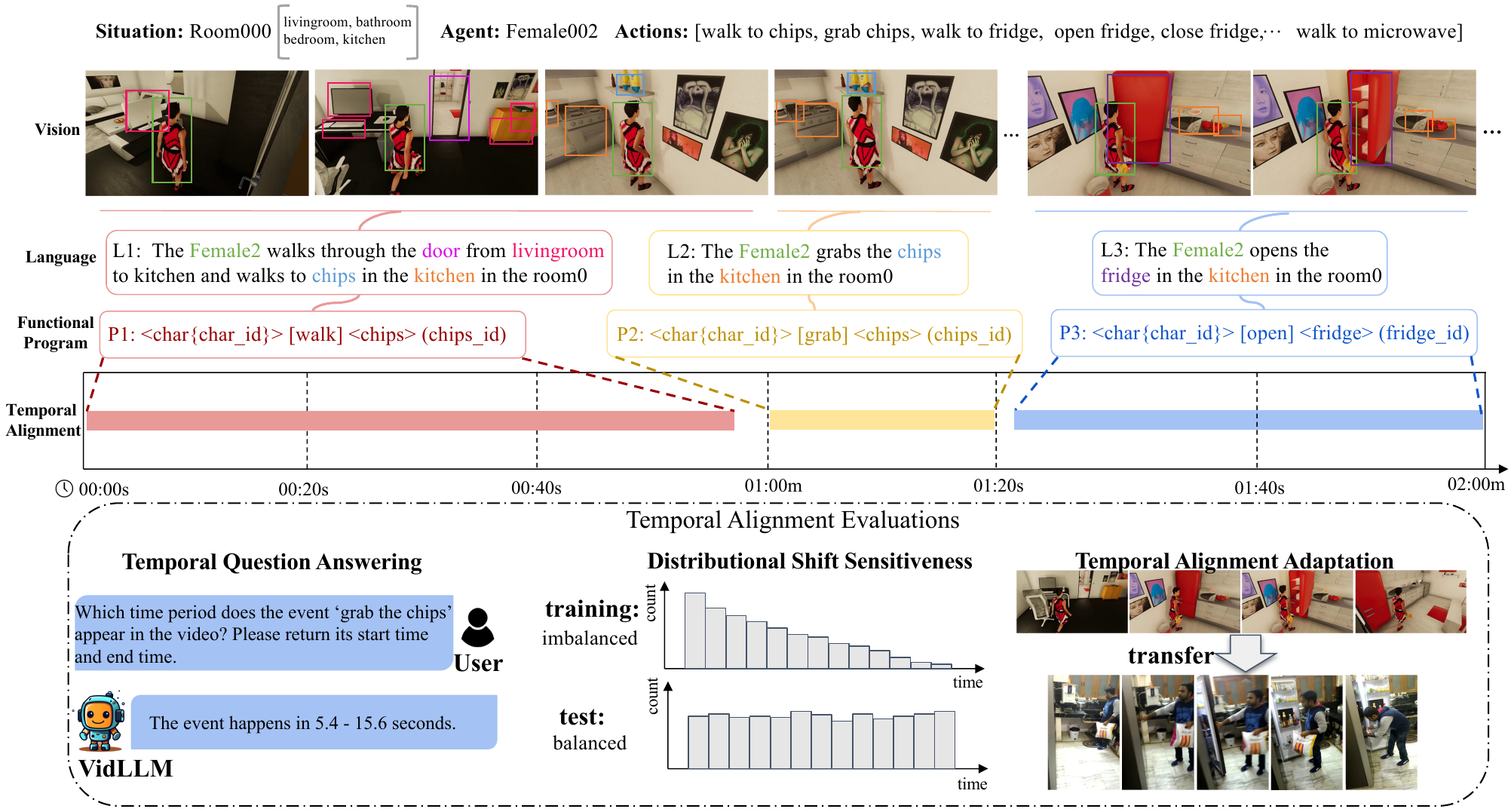}
    \centering
    \caption{Overview of the 
    SVLTA benchmark, which consists of synthetic videos, language, and high-quality temporal alignment.}
    \label{fig:svlta_overview}
    \vspace{-5mm}
\end{figure*}

To address similar goals, one category of existing work builds datasets for traditional video grounding models on top of collected videos and temporal segment annotations, such as TACoS, DiDeMo, and Charades-STA~\cite{gao2017tall,krishna2017dense,anne2017localizing,regneri2013grounding,soldan2022mad}. But human-crafted annotations solely on semantic correlations inevitably result in unreliable labels~\cite{otani2020uncovering,zhou2021embracing}, due to the inherent subjectivity and ambiguity of linguistic and visual semantic descriptions. Additionally, while constructing data combinations with semantic correlation as the goal allows for consideration of semantic diversity, it has been found to suffer from significant temporal distribution imbalances~\cite{otani2020uncovering,yuan2021closer}. As multimodal large model capabilities advance, these shortcomings in evaluations become critical bottlenecks affecting the accuracy and objectivity of evaluations.
We conduct a comprehensive examination of the current video benchmarks in depth. We identify three types of temporal alignments (processes, compositions, and entities) from a decomposed perspective, propose the metric Temporal Jenson-Shannon Divergence (TJSD), and visualize the temporal distributions for clearer understanding. The observed biased distributions at multiple levels reflect the ``unbalanced influences'' of the existing evaluations.

Distinct from previous work, we propose SVLTA, exploring the synthetic vision-language temporal alignment, enabling compositional, unbiased, and large-scale video evaluations with precise temporal alignments. SVLTA encompasses 96 different compositional actions, 25.3K synthetic video situations, and 77.1K high-quality temporal annotations with consistent visual-language semantics.
Our benchmark addresses the limitations of existing datasets by generating videos through synthetic simulations, providing better control over temporal alignment, which is challenging in realistic videos. The benchmark overview is illustrated in Figure~\ref{fig:svlta_overview}. 
Synthetic video situations are created by executing functional programs derived from a series of raw data (predefined agents, actions, and situations) in a human-centric 3D VirtualHome simulator~\cite{puig2018virtualhome,liao2019synthesizing,puig2021watchandhelp}, while sentences are generated using templates defined for different scenes. 
For timestamp annotations, the programs automatically record the time and duration of each action during execution. This method allows us to easily associate the annotated timestamps with the corresponding actions in the sentences, resulting in high-quality annotations.

Through SVLTA, we evaluate temporal alignment from three perspectives: 1) temporal question answering, 2) distributional shift sensitiveness, and 3) temporal alignment adaptation. Our experiments yield the following results: 1) by using simple temporal-related questions, current popular Video Large Language Models (VidLLMs) rarely give correct answers, even some time-aware VidLLMs or close-sourced models, which means current models lack temporal alignment capability, 2) most specific temporal alignment models are easily affected by temporal bias, even some debiased models, which indicates these models have poor generalization and cannot address temporal distribution shifts, and 3) several specific temporal alignment models may have potential to transfer temporal knowledge, which means these models can adapt to new situations in some degree.
Our main contributions are as follows:
\begin{itemize}
    \item We conduct three levels of temporal distribution analysis and visualization from a decomposed perspective and propose an appropriate measurement.
    \item We introduce SVLTA, a synthetic benchmark for vision-language temporal alignment. Through our proposed approaches, we generate both human activity situations, language descriptions, and temporally aligned samples with fewer human resources based on commonsense inference, compositional optimization, and constrained sampling. 
    \item We design multiple types of vision-language temporal alignment tasks enabling comprehensive evaluations for both pre-trained VidLLMs and specific temporal alignment models, providing detailed experimental analysis with insightful conclusions.
\end{itemize}
\section{Related Works}
\label{sec:related}

\subsection{Visual-Language Temporal Alignment}

Visual-language temporal alignment aims to link video content and language in the temporal dimension, which also appears in action localization or temporal grounding tasks. Most of the benchmarks in temporal grounding or localization~\cite{gao2017tall,anne2017localizing,regneri2013grounding,lei2020tvr,grauman2022ego4d,song2023egod} collected videos from web or recorders and utilized the crowd-sources to annotate temporal segments (e.g. action timestamps, event order, etc).
Additionally, some of them~\cite{soldan2022mad,han2022temporal} used Automatic Speech Recognition (ASR) to generate text transcripts from speech and constructed their related timestamps which require less costs and resources. 
These benchmarks played an important role in evaluating the development of temporal alignment systems~\cite{chen2018temporally,yuan2019find,yuan2019semantic,zeng2020dense,lei2021detecting,lin2023univtg,huang2024vtimellm,li2024lego}, which pushes the boundaries of temporal alignment research. 
According to our analysis (refer to \S~\ref{temporal_analysis}), current video benchmarks may be influenced by multiple levels of bias as they are primarily sourced from the real world and rely on human-provided annotations. Our conclusion also consists of the observations of several recent work~\cite{otani2020uncovering,yuan2021closer,2024videotree}, and they noticed serious temporal biases of sentence-level video segments in video datasets, which can cause the model to have poor generalization when training on these datasets~\cite{nan2021interventional,liu2022reducing,lan2023curriculum}. 
Moreover, Otani et al.~\cite{otani2020uncovering} found that the temporal annotations from multiple annotators are inconsistent since they have different perceptions that lead to unreliable annotations. 
These drawbacks limit the accuracy and effectiveness of the assessment, which demonstrates current benchmarks cannot provide a detailed and valid diagnosis environment for temporal alignment models. 

\subsection{Synthetic Situation Generation}
Synthetic data generation is gaining significant interest within the research community because of its cost-effectiveness and ease of scale, with applicability across various research fields. Synthetic videos were widely adopted in data augmentation for video understanding or action recognition~\cite{roberto2017procedural,varol2021synthetic,hwang2021eldersim,guo2022learning,chang2023learning,kim2022transferable,qiu2023virtualhome}. Meanwhile, some research~\cite{Girdhar2020CATER,hazra2023egotv, Yi2020CLEVRER} utilized the templates or physics engine to generate associated questions for diagnostic evaluations on model capacities in neighbor tasks, such as video question answering~\cite{Yi2020CLEVRER}, action recognition~\cite{kim2022transferable, Girdhar2020CATER}, and multi-object tracking~\cite{gaidon2016virtual}. However, the aforementioned methods are not designed to study vision-language temporal alignment and ignore the explicit control of temporal alignment as the primary generation objective. Unlike the others, our SVLTA generates human activity situations, language descriptions, and temporally aligned samples for multiple types of program-generated evaluation tasks based on commonsense knowledge and compositional optimization. 

\subsection{Temporal Understanding}

With the emergence of Video Large Language Models, there has been a significant increase in the collection of video-language benchmarks, to evaluate them. Most previous work mainly considered semantic diversity and video length duration when collecting the data, while the video has a temporal dimension that depicts the objects' motion and the corresponding interaction state. To fill this blank and comprehensively evaluate the video-language models, some works aim to explore the temporal understanding ability of these multimodal video models. AGQA~\cite{grunde2021agqa} utilized templates combined with the spatio-temporal scene graph to generate well-designed questions and answers to assess the temporal reasoning ability. Furthermore, ViLMA~\cite{kesen2024vilma} and Perception Test~\cite{patraucean2023perception} developed a series of temporal-related tasks (for example, action location and action counting) to diagnose whether the model has strong temporal modeling capabilities. TempCompass~\cite{liu2024tempcompass} contributed a comprehensive benchmark to evaluate temporal perception using the single frame, language debiasing strategies, and various task formats. E.T.Bench~\cite{liu2024bench} also designed multiple fine-grained temporal sensitive tasks to assess event-level video understanding from open-ended scenarios. However, none of the above benchmarks focus on the assessment of temporal alignment with fairness and comprehensiveness, which is also an important part of temporal understanding.
\section{SVLTA}
\label{sec:method}

SVLTA is a synthetic and scalable benchmark with diverse, compositional, and controllable temporal distribution, to provide a fair diagnostic framework for evaluating the temporal alignment ability of models.
However, constructing such a benchmark is challenging, especially in controlling the temporal distribution, as it requires detailed types of temporal distribution and carefully designed methods to maintain the balance of the data set.

In this work, we first formulate the visual-language temporal alignment problem (\S~\ref{problem_formulation}), then thoroughly analyze the temporal distribution in current mainstream datasets and design a metric to measure them (\S~\ref{temporal_analysis}). Following this, common sense activity and action chain generation, controllable temporal distribution strategies, and synthetic generation are adopted to build a benchmark that contains three processes: 1) synthetic video generation (\S~\ref{video_gen}), 2) language sentence generation (\S~\ref{sentence_gen}), and 3) visual-language temporal alignment (\S~\ref{timestamp_gen}), a post-processing filtering method is also developed to further adjust the temporal distribution (\S~\ref{ICGF}). Finally, we compare our SVLTA with other major benchmarks (\S~\ref{benchmark_comp}).

As summarized in Table~\ref{tab:overview_comparison}, SVLTA comprises 25.3K dynamic situations derived from human activity videos, featuring 77.1K language descriptions and temporal-aligned activity sequences, covering 96 distinct compositional actions. The benchmark provides 77.1K high-quality temporal alignment annotations, with average video and moment durations of 134.1 and 24.3 seconds, respectively. Up to this point, it is a novel benchmark with compositional, controllable, and unbiased temporal distributions.

\begin{table*}[t]
\footnotesize
\caption{Comparison of SVLTA and existing benchmarks for Vision-Language Temporal Alignment. SVLTA is a synthetic benchmark with controllable, compositional, and unbiased data distributions. N/A: not available.}
\vspace{-5mm}
\label{tab:overview_comparison}
\begin{center}
\begin{tabular}{@{}ccccccccc@{}}
\toprule
& \multicolumn{3}{c}{\textbf{Dataset Statistics}} & \multicolumn{5}{c}{\textbf{Dataset Characteristics}} \\
\cmidrule(lr){2-4} \cmidrule(lr){5-9}
\textbf{Benchmark}  & \makecell[c]{\# Videos / \\ \# Annotations} & \# Actions  & \makecell[c]{Avg. Video / Moment \\ Duration (s)} & Scalable & Controllable & Synthetic & Compositional  & Unbiased  \\
\midrule
TACoS~\cite{regneri2013grounding} & 0.1K / 18.8K & 60 & 287.1 / 27.9 & \textcolor{red}{\ding{55}} &  \textcolor{red}{\ding{55}} & \textcolor{red}{\ding{55}}  & \textcolor{red}{\ding{55}}  &  \textcolor{red}{\ding{55}}  \\ 
ActivityNet Captions~\cite{krishna2017dense} & 14.9K / 54.9K  & N/A &  117.6 / 37.1 &  \textcolor{red}{\ding{55}} &  \textcolor{red}{\ding{55}}  &  \textcolor{red}{\ding{55}} &  \textcolor{red}{\ding{55}}  & \textcolor{red}{\ding{55}} \\
Charades-STA~\cite{gao2017tall} & 6.7K / 16.1K  & 157 & 30.0 / 8.1  &  \textcolor{red}{\ding{55}}  &  \textcolor{red}{\ding{55}}  &  \textcolor{red}{\ding{55}}   &  \textcolor{red}{\ding{55}} & \textcolor{red}{\ding{55}}  \\
DiDeMo~\cite{anne2017localizing} & 10.5K / 40.5K  & N/A & 30.0 / 6.5 &  \textcolor{red}{\ding{55}} &  \textcolor{red}{\ding{55}}  & \textcolor{red}{\ding{55}}  &  \textcolor{red}{\ding{55}}  &  \textcolor{red}{\ding{55}} \\
TVR~\cite{lei2020tvr} & 21.8K / 109K  &  N/A & 76.1 / 9.1  &  \textcolor{red}{\ding{55}} & \textcolor{red}{\ding{55}}  &  \textcolor{red}{\ding{55}}  & \textcolor{red}{\ding{55}}  & \textcolor{red}{\ding{55}} \\
MAD~\cite{soldan2022mad} &  0.7K / 384.6K & N/A &  6646.2 / 4.1 & \textcolor{green}{\ding{52}}  &  \textcolor{red}{\ding{55}}  &  \textcolor{red}{\ding{55}}  &  \textcolor{red}{\ding{55}} &  \textcolor{red}{\ding{55}} \\
Ego4D~\cite{grauman2022ego4d} & 1.6K / 19.2K &  N/A &  495.3 / 11.2 &  \textcolor{red}{\ding{55}}  & \textcolor{red}{\ding{55}}  &  \textcolor{red}{\ding{55}} &  \textcolor{red}{\ding{55}}  &  \textcolor{red}{\ding{55}} \\
Ego4D Goal-Step~\cite{song2023egod}  & 0.8K / 48K  & N/A & 1560.0 / 32.5 & \textcolor{red}{\ding{55}} &  \textcolor{red}{\ding{55}} & \textcolor{red}{\ding{55}} & \textcolor{red}{\ding{55}} & \textcolor{red}{\ding{55}} \\
E.T.Bench~\cite{liu2024bench} & 7K / 7.3K & N/A & 129.0 / 11.0 & \textcolor{red}{\ding{55}} & \textcolor{red}{\ding{55}} & \textcolor{red}{\ding{55}} & \textcolor{red}{\ding{55}} & \textcolor{red}{\ding{55}} \\
\midrule
SVLTA (ours) & 25.3K / 77.1K &  96 & 134.1 / 24.3 & \textcolor{green}{\ding{52}}  & \textcolor{green}{\ding{52}} & \textcolor{green}{\ding{52}} & \textcolor{green}{\ding{52}} & \textcolor{green}{\ding{52}} \\
\bottomrule
\end{tabular}    
\end{center}
\vspace{-7mm}
\end{table*}

\subsection{Problem Formulation}
\label{problem_formulation}

We define visual-language temporal alignment as the task of synchronizing video and language in the temporal domain, aiming to identify the timestamps of video moments that most closely match the semantics of the corresponding sentences. In particular, we denote an untrimmed video as $V=\left\{ v_i \right\}_{i=1}^{M}$ and a sentence in the language as $L=\left\{ l_j \right\}_{j=1}^{N}$, where $v_i$ is a frame in the video and $l_j$ means a word in the sentence. The goal of the visual-language temporal alignment task is to build a model $f$ with the input $V$ and $L$ that can correctly predict the start time $t_s$ and the end time $t_e$ of the moment, which are formulated as follows:
\begin{equation}
    [t_s, t_e] = f(V, L; \theta)
\end{equation} 
where $M$ and $N$ are respective length of video and text, and $\theta$ is the model parameter.

\subsection{Temporal Distribution Analysis}
\label{temporal_analysis}

\begin{figure}[t]
    \includegraphics[width=1.0\linewidth]{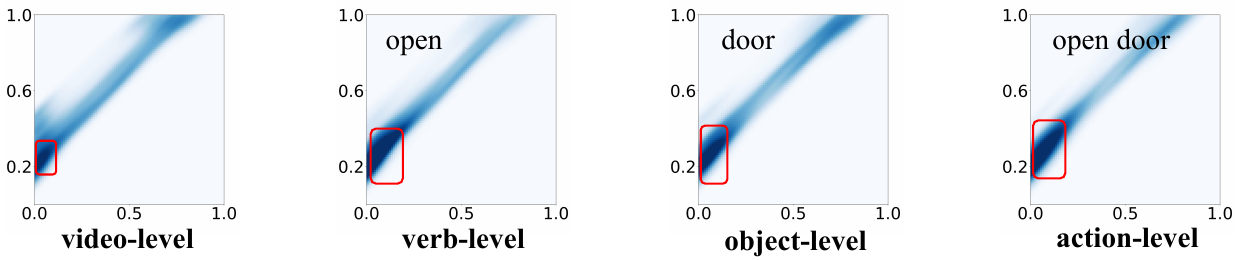}
    \centering
    \caption{Multiple Levels of Temporal Distributions. We sample decomposed semantic constituents in the Charades-STA. The color darkness represents the sample density. The horizontal and vertical axes represent the normalized start and end time points.}
    \label{fig:temporal_bias_vis}
    \vspace{-6mm}
\end{figure}

\noindent \textbf{Temporal Distribution in Decomposition Perspective} 
We first explore multiple levels of temporal alignments in the existing video benchmarks and valid if they are appropriate to the vision-language temporal alignment evaluations. Inspired by several works~\cite{chen2020fine,li2022compositional}, we take a decomposition perspective and treat a situation as comprising actions, with each action containing a verb-noun structure. Thus, the semantic constituents and video segments are associated on multiple levels. 
As illustrated in Figure~\ref{fig:temporal_bias_vis}, the visualizations reveal that temporal distributions are influenced by several biases, ranging from global to local levels. The \textbf{process temporal bias (video-level)} indicates limitations in overall data selection, while the \textbf{composition temporal bias (action-level)} and \textbf{entity temporal bias (verb/object-level)} result in evaluations focusing only on particular positions of temporal segments in videos, neglecting others.

\noindent \textbf{Quantitative Comparison via Temporal Divergence} 
To effectively analyze the imbalance of temporal distributions within datasets, quantifiable metrics are necessary. 
Drawing inspiration from prior research~\cite{collins2018evolutionary,xiang2020learning} that designed metrics to measure class imbalance in classification benchmarks, we propose a new metric, Temporal Jensen–Shannon Divergence (TJSD), to measure the differences between the target distribution and the ideal distribution. The target distribution means the temporal distribution of the current dataset and the ideal distribution denotes the uniform distribution. To address the problem that time is continuous without natural categorization, we first divide the video into $n$ equal moments to discrete time, leading to $\frac{n(n+1)}{2}$ different temporal bins, each bin represents a temporal class,  and then we assign the timestamps into these bins. Therefore, the target temporal distribution can be represented by the number of samples in these bins and the ideal distribution means that the number of samples in each bin is the same. Finally, the Jensen–Shannon divergence is utilized to calculate the difference between the target and uniform temporal distribution. The detailed TJSD equation is in \emph{Supplementary}. The statistics of existing datasets are shown in Table~\ref{tab:temporal_biases_comp}, demonstrating that although some benchmarks have smaller process temporal bias, they ignore other types of temporal bias when collecting the videos.


\begin{figure*}[t]
    \includegraphics[width=0.9\linewidth]{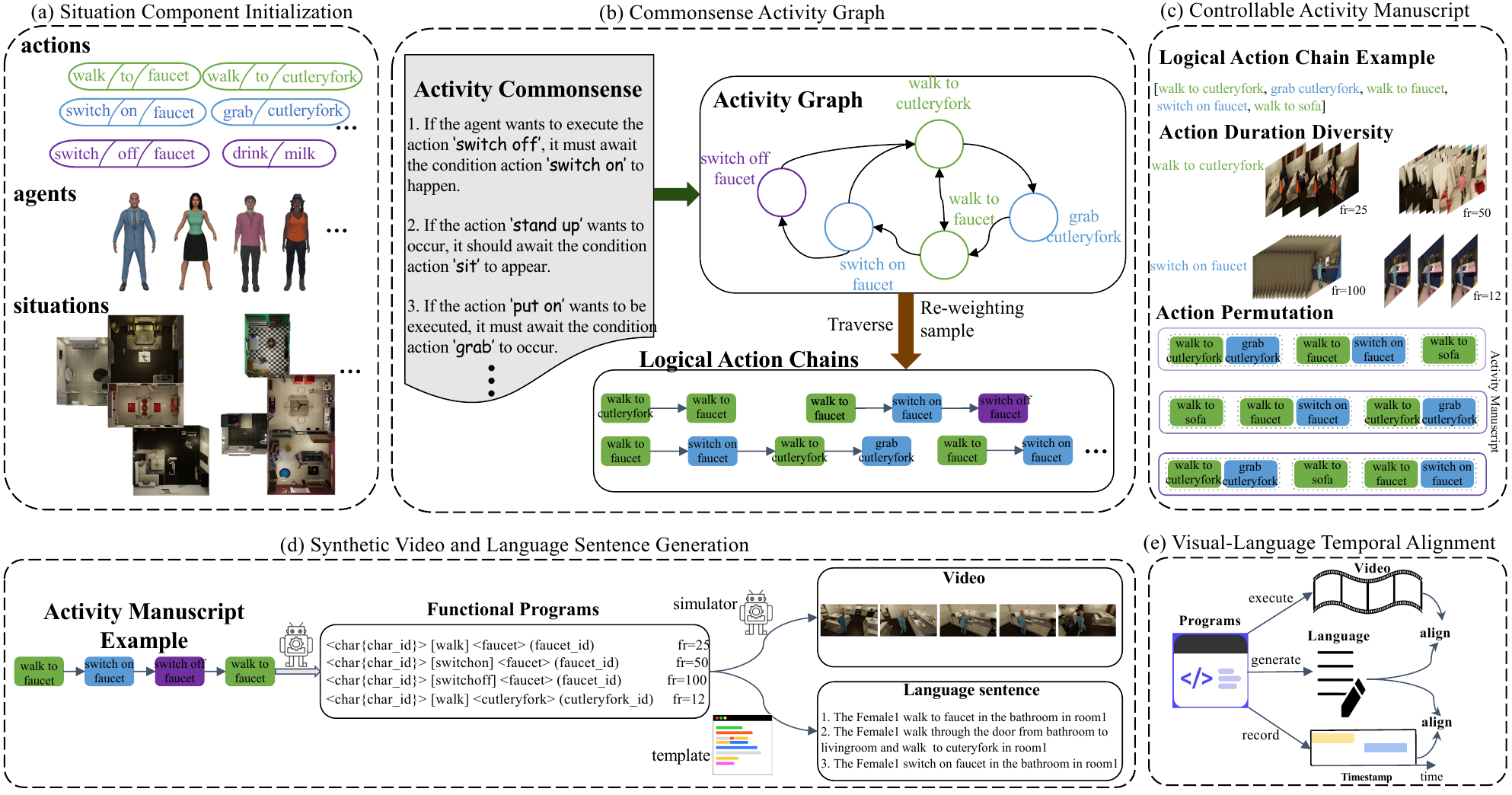}
    \centering
    \caption{Overview of the benchmark generation process, which contains (a): Situation Component Initialization defines a series of compositional elements, which includes diverse actions, agents, and situations, (b): Commonsense Activity Graph builds a graph on the activity commonsense and then use the traversal algorithm and re-weighting sampling to acquire various and meaningful logical action chains, (c): Controllable Activity Manuscript operates the actions in logical action chains through different framerates and permutations to obtain the final activity manuscript, thereby balancing the temporal distribution, (d): Synthetic Video and Language Sentence Generation convert the generated activity manuscript to the functional programs and utilize it to generate synthetic videos and sentences, and (e): Visual-Language Temporal Alignment automatically associates the timestamps with the action in the sentence to obtain high-quality annotations.}
    \label{fig:svlta_gen}
    \vspace{-6mm}
\end{figure*}

\subsection{Benchmark Generation}
\subsubsection{Synthetic Situation Generation}
\label{video_gen}

\noindent \textbf{Situation Component Initialization} 
To create a temporal alignment dataset with high-quality and diverse temporal alignment samples, we generate the synthetic videos by defining compositional situation components and functional programs via a human-centric simulator VirtualHome~\cite{puig2018virtualhome,liao2019synthesizing,puig2021watchandhelp}. 
To begin with, we establish the diverse action, situation, and agent spaces for video generation, as shown in Figure~\ref{fig:svlta_gen} (a). We select 96 meaningful actions that can be executed within the simulator, 7 scenes that serve as the environments for these actions, and 6 alternative agents who act as characters to carry out the pre-defined actions.
Executing functional programs will trigger an agent to perform the pre-defined activity in a virtual home scene.

\noindent \textbf{Commonsense Activity Graph} After initializing component spaces, a conventional strategy is to generate synthetic activity videos by selecting several actions. However, random selection would lead to meaningless combinations of actions, making it challenging to accomplish our objective. 
VirtualHome provides basic logical rules between paired actions, i.e. some actions must wait until the conditional actions are completed before they can happen. But longer action sequences would encounter unreasonable compositions accidentally.
For instance, if four actions \emph{walk to fridge}, \emph{open fridge}, \emph{close fridge}, and \emph{grab sandwich} are selected, only the first three actions can be performed correctly and the last one is meaningless since \emph{grab sandwich} and \emph{walk to fridge} are inconsistent.
Therefore, we manually check the defined rules and only keep the reasonable and executable action relations and agents in situations, which consistent with human commonsense knowledge in the real world.
Based on this, we can get all potential relationships between these actions, and an activity graph is built upon to generate diverse and reasonable action compositions. 
Several recent works adopt novel ideas to enhance the generation quality~\cite{sok-bench2024, wu2021star,ji2020action}. Our activity graph inspired from them and specialized in following aspects: 1) our activity graph is built on commonsense knowledge with pre-defined action sets while others design the graph based on real-world videos, 2) our graph aims to generate new synthetic videos yet others are used to produce novel questions, answers, and annotations. 
Then, we adopt the graph traversal algorithm (DFS~\cite{cormen2022introduction} or BFS~\cite{cormen2022introduction}) to generate logical action chains by traveling action nodes in activity graphs with given lengths.
However there are different levels of constraint between the actions in the activity commonsense, i.e., some actions have fewer conditional actions and they do not need to wait for other actions to happen, thus the node degree is imbalanced in the activity graph. To solve this problem, a re-weighting sampling strategy is proposed to ensure that all candidate actions have a uniform probability of being selected in each traversal.
The complete process is illustrated in Figure~\ref{fig:svlta_gen} (b) and the strategy details are in \emph{Supplementary}.

\noindent \textbf{Controllable Activity Manuscript} Directly utilizing logical action chains to generate videos can introduce potential temporal bias since the action positions and durations are uncontrollable. Previous work~\cite{soldan2022mad} attempted to mitigate process imbalances in temporal distribution by collecting long-term videos. However, this approach fails to address other types of temporal bias within the video and does not provide essential solutions to this critical issue.
We propose two strategies to produce better temporal distribution by controlling the positions and durations of actions: Action Duration Diversity (ADD) and Action Permutation (AP). As shown in Figure~\ref{fig:svlta_gen} (c), the idea is to ensure each action can appear at any position in the video and have diverse durations. 
Specifically, AP permutates the actions in chains so that each action will appear in as many positions as possible while satisfying the activity commonsense. ADD enables the diverse action durations by adopting varied video framerates. We employ ADD and AP to create controllable activity manuscripts for generating synthetic videos.

\noindent \textbf{Synthetic Video Generation} We randomly choose an agent and a situation from their corresponding spaces. Then we execute functional programs with the activity manuscript, agent, and situation in the simulator to create situation videos, as shown in Figure~\ref{fig:svlta_gen} (d).

\subsubsection{Language Sentence Generation}
\label{sentence_gen}

For text generation, previous works~\cite{gao2017tall,krishna2017dense,lei2020tvr,soldan2022mad,han2022temporal} produced the language sentence by crowdsourcing annotation or an ASR model. These methods may have two drawbacks: 1) ambiguity problem, since different human annotators would write different semantic text for a sample, this problem is also proposed in the \cite{zhou2021embracing}, and 2) noise issue, pre-trained models sometimes provide unreliable results, which leads to incorrect generated text. These disadvantages would reduce the quality of the benchmark and increase the challenges of model training. Thus, we utilize template-based generation to create the sentence to get high-quality language queries. In detail, three templates are defined to directly convert each action in action chains into sentences with different scenes and agents based on whether the scenes change when the action occurs, as shown in \emph{Supplementary}. We utilize all actions that happened in the videos and use templates directly to construct the sentences, which can reduce the ambiguity and noise problems in the dataset. Additionally, since the Large Language Models demonstrate superior performance in natural language generation, we use the GPT-3.5-turbo to rewrite the original template-based sentences into more natural and diverse descriptions, serving as an auxiliary resource to strengthen our benchmark. This process is exhibited in Figure~\ref{fig:svlta_gen} (d).

\subsubsection{Vision-Language Temporal Alignment}
\label{timestamp_gen}

After we derive the synthetic videos and languages by prior steps, we need to align them to produce corresponding timestamps. Previous works~\cite{gao2017tall,krishna2017dense,lei2020tvr} let humans assign each language query to the video content, but it may cause noisy temporal labels as mentioned in \cite{otani2020uncovering}. Thanks to the VirtualHome, when we generate synthetic videos, it records the time and duration of each action. Consequently, we only need to associate the automatically annotated timestamp with the action in the sentence to generate high-quality temporal annotations, which is depicted in Figure~\ref{fig:svlta_gen} (e).

\begin{table}[t]
\footnotesize
\caption{Comparison of multi-level temporal biases.}
\vspace{-6mm}
\label{tab:temporal_biases_comp}
\begin{center}
\setlength{\tabcolsep}{5pt}
\begin{tabular}{@{}ccccc@{}}
\toprule
 &  & \multicolumn{2}{c}{\textbf{Entity}} &  \\
\cmidrule(lr){3-4}
\textbf{Benchmark} & \textbf{Process} & Verb & Object &  \textbf{Composition} \\
\midrule
TACoS~\cite{regneri2013grounding} & 0.243 & 0.786 & 0.787 & 0.899 \\
ActivityNet Captions~\cite{krishna2017dense} & 0.107 & 0.764 & 0.827 & 0.921 \\
Charades-STA~\cite{gao2017tall} & 0.287 & 0.739 & 0.877 & 0.881 \\
TVR~\cite{lei2020tvr} & 0.229 & 0.779 & 0.84 & 0.914 \\
MAD~\cite{soldan2022mad} & 0.628 & 0.842 & 0.869 & 0.926 \\
\midrule
SVLTA (ours)  & \textbf{0.073} & \textbf{0.266} & \textbf{0.101} & \textbf{0.322} \\ 
\bottomrule
\end{tabular}    
\end{center}
\vspace{-8mm}
\end{table}

\subsubsection{Inequality Constrained Global Filtering}
\label{ICGF}

Though we propose two strategies to control the temporal distribution, they only operate the local distribution in each logical action chain, which may produce potential temporal biases from the global perspective. Therefore, a debiasing method should be utilized to balance the temporal distributions as a post-processing step. 
Previous research~\cite{sakaguchi2020winogrande,le2020adversarial} designed the Adversarial Filtering (AF) method to reduce the bias in the dataset to achieve the balanced distribution goal. However, AF only has sub-optimal results since it filters those samples that most influence the distribution in each iteration. To perform a better debiasing effect, we propose a novel approach Inequality Constrained Global Filtering (ICGF) to adjust the temporal distribution of each action since a video is composed of multiple different actions. The main idea of ICGF is to filter some samples to obtain a more balanced temporal distribution while not filtering too many samples. Specifically, we treat this idea as a nonlinear optimization problem with inequality constraints, the optimization goal is to reduce the gap between the current distribution and the uniform distribution (we use an absolute deviation function to measure the distribution gap) and the constraint is that too many samples should not be filtered (a filtering rate is utilized to control sample size). The details of ICGF and its comparison with other methods are included in \emph{Supplementary}.

\subsubsection{Benchmark Comparison}
\label{benchmark_comp}
\begin{figure}[t]
    \includegraphics[width=1.0\linewidth]{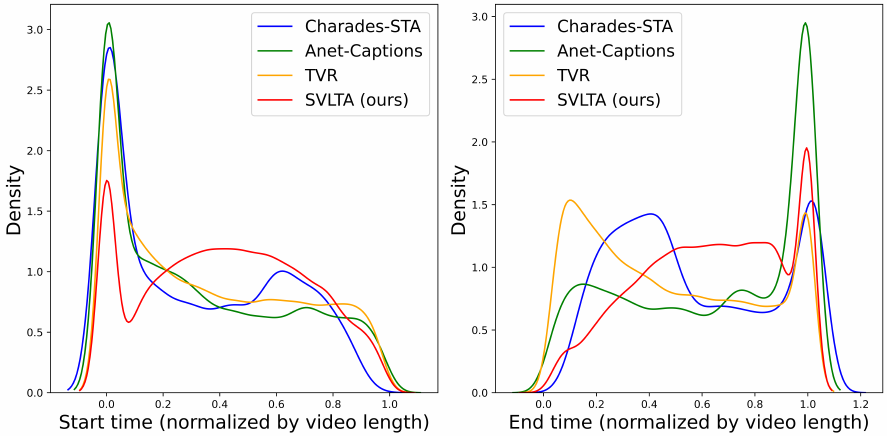}
    \centering
    \caption{Temporal distributions of beginning or ending times.}
    \label{fig:temporal_dist_comp}
    \vspace{-6mm}
\end{figure}

To evaluate our strategies for balancing temporal distributions and the effectiveness of our data-level debiasing approach ICGF, we first compare the temporal bias in our SVLTA dataset with five mainstream datasets: MAD~\cite{soldan2022mad}, TACoS~\cite{regneri2013grounding}, ActivityNet Captions (Anet-Captions)~\cite{krishna2017dense}, Charades-STA~\cite{gao2017tall}, and TVR~\cite{lei2020tvr}. MAD features long-term videos (over 1 hour), while TACoS, Anet-Captions, and TVR contain medium-term videos (over 1 minute), and Charades-STA includes short-term videos (around 30 seconds). The diverse characteristics of these datasets enable a comprehensive comparison, as shown in Table~\ref{tab:temporal_biases_comp}. Our results indicate that SVLTA exhibits the least temporal bias across various metrics, highlighting the effectiveness of our synthetic generation method in creating well-controlled temporal alignments. 
Additionally, we plot the temporal distribution of the moment start and end times for all temporal annotations, as illustrated in Figure~\ref{fig:temporal_dist_comp}. The results show that the distribution curve of SVLTA looks flatter and has a smaller variance than other datasets, indicating the validity of our controllable strategies and filtering methods. 

\section{Experiment}

We diagnose the temporal alignment ability of models from three perspectives: 1) temporal question answering, which aims to evaluate the temporal alignment of current VidLLMs, 2) distributional shift sensitiveness attempts to analyze the impact of temporal distribution shift on temporal alignment, and 3) temporal alignment adaptation explores whether the temporal alignment capability of models can be transferred to the new video situations or domains. 

\subsection{Experimental Setting}
\label{sec:resul}
Temporal question answering utilizes the simple question-answer paradigm, which is widely used in other benchmarks~\cite{li2023seed,yue2023mmmu,liu2023mmbench} that evaluate the comprehension and reasoning ability of MLLMs. Here, we utilize the same temporal-related question prompt as used in the previous works~\cite{ren2024timechat,liu2024bench}. Additionally, Charades-STA is selected as the target for new video situations transferring in the temporal alignment adaptation since Charades-STA is collected in the real-world home, which has situations and actions similar to our benchmark. In distributional shift sensitiveness, a set of training, validation, and test sets are initially created with high temporal bias through long-tailed sampling, then a low temporal bias test set is constructed using ICGF from the remaining samples, and finally evaluating the difference between the results of the two test sets.

\subsection{Evaluation Model} 

We evaluate both VidLLMs and specific temporal alignment models. For the temporal question answering, we analyze several VidLLMs, including open-sourced models, such as Video-LLaMA2~\cite{cheng2024videollama}, Video-ChatGPT~\cite{maaz2023video}, Video-LLaVA~\cite{lin2023video}, Videochat2~\cite{li2024mvbench}, and LLaVA-Video~\cite{zhang2024video}, as well as close-sourced models, such as Gemini 1.5 Pro~\cite{team2024gemini}, GPT-4o~\cite{GPT-4o}, and GPT-4o-mini~\cite{GPT-4o}. Additionally, time-aware VidLLMs like TimeChat~\cite{ren2024timechat}, VTimeLLM~\cite{huang2024vtimellm}, and E.T.Chat~\cite{liu2024bench} are also considered.  For distributional shift sensitiveness and temporal alignment adaptation, we benchmark different specific frameworks: 1) anchor-free: VSLNet~\cite{zhang2020span} and LGI~\cite{mun2020local}, 2) anchor-based: 2D-TAN~\cite{zhang2020learning}, and 3) transformer-based: QD-DETR~\cite{moon2023query}. Meanwhile, two debiased models DCM~\cite{yang2021deconfounded} and Shuffling~\cite{hao2022can} are also diagnosed for distributional shift sensitiveness. 


\subsection{Evaluation Metric} 
For temporal question answering and temporal alignment adaptation, we employ the same metrics as used in prior studies~\cite{gao2017tall,yuan2019find} to show their performance, namely $R@1, IoU=0.1, 0.3, 0.5, 0.7, 0.9$ and mean IoU (mIoU). Furthermore, a new metric RC is developed to assess distributional shift sensitiveness. The motivation of RC is that a temporal robust model should not be easily affected by temporal bias when training, meaning it could perform reliably on test sets with varying distributions. In our setting, the RC is the difference between the results of the two test sets when training on the high temporal bias data. The higher the RC, the worse the temporal robustness of the model. 

\begin{table*}[h]
\footnotesize
\caption{The results of current popular open-sourced and close-sourced VidLLMs on SVLTA.}
\vspace{-6mm}
\label{tab:vidllms_comp}
\begin{center}
\setlength{\tabcolsep}{8.4pt}
\begin{tabular}{@{}cccccccccc@{}}
\toprule
&  &  & & &  \multicolumn{4}{c}{\textbf{R@1}} &  \\
\cmidrule(lr){6-9}
\textbf{Method} & \textbf{\# Frames} & \textbf{Size} & \textbf{Visual Encoder} & \textbf{LLM} & IoU=0.1 & IoU=0.3 & IoU=0.5 & IoU=0.7 & \textbf{mIoU} \\
\midrule
\multicolumn{10}{c}{\textcolor{gray}{General Open-sourced Models: All models use their default setting. Except LLaVA-Video, due to the GPU memory limits.}} \\
\midrule
LLaVA-Video~\cite{zhang2024video} & 16 & 7B & SIGLIP-SO400M & Qwen2 & 2.52 & 0.89 & 0.40 & 0.27 & 0.84  \\
Videochat2~\cite{li2024mvbench} & 16 & 7B & UMT-L/16 & Vicuna-0 &  2.93 & 0.87 & 0.32 & 0.13 & 0.87 \\
Video-LLaVA~\cite{lin2023video} & 8 & 7B &  \fontsize{6pt}{6pt}\selectfont{LanguageBind-ViT-L/14} & Vicuna-1.5 & 8.22 & 3.19 & 0.96 & 0.23 & 2.59 \\
Video-ChatGPT~\cite{maaz2023video} & 100 & 7B & CLIP-ViT-L/14 & Vicuna-1.1 & 10.68 & 3.17 & 0.90 & 0.21 & 2.94 \\
Video-LLaMA2~\cite{cheng2024videollama} & 16 & 7B & CLIP-ViT-L/14 & Mistral-7B & 35.48 & 16.02 & 6.64 & 2.28 & 12.33 \\
\midrule
\multicolumn{10}{c}{\textcolor{gray}{Time-aware Open-sourced Models: All models utilize their default configuration.}} \\
\midrule
E.T.Chat~\cite{liu2024bench} & 1FPS & 3.8B &  EVA-ViT-G/14 & Phi-3-Mini & 17.86 & 8.07 & 3.48 & 1.36 & 6.29 \\
TimeChat~\cite{ren2024timechat} & 96 & 7B & EVA-ViT-G/14 & Llama-2 &  23.29 & 13.58 & 6.96 & 3.25 & 9.61 \\
VTimeLLM~\cite{huang2024vtimellm} & 100 & 7B & CLIP-ViT-L/14  & Vicuna-1.5 & 29.97 & 13.29 & 5.26 & 1.71 & 10.29 \\
\midrule 
\multicolumn{10}{c}{\textcolor{gray}{Close-sourced Models: Evaluated on a subset with 2000 samples.}} \\
\midrule
GPT-4o-mini~\cite{GPT-4o} & 32 &  --- & --- & --- & 24.79 & 6.49 & 1.57 & 0.42 & 6.70\\
Gemini 1.5 Pro~\cite{team2024gemini} & 1FPS & --- & --- & --- & 32.30 & 17.45 & 7.45 & 3.15 & 12.48 \\
GPT-4o~\cite{GPT-4o} & 32 & --- & --- & --- & 49.54 & 27.38 & 11.69 & 5.62 & 18.90  \\
\bottomrule
\end{tabular}
\end{center}
\vspace{-5mm}
\end{table*}

\begin{table*}[h]
\footnotesize
\caption{The performance of distributional shift sensitiveness on SVLTA.}
\vspace{-6mm}
\label{tab:temporal_bias_sensitive}
\begin{center}
\setlength{\tabcolsep}{13pt}
\begin{tabular}{cccS[table-format=2.2]S[table-format=2.2]S[table-format=2.2]S[table-format=2.2]S[table-format=2.2]c}
\toprule
&  &  & \multicolumn{4}{c}{\textbf{R@1}} &  &  \\
\cmidrule(lr){4-7} 
& \textbf{Method}  & \textbf{Test set} & {IoU=0.3}  & {IoU=0.5}  & {IoU=0.7} & {IoU=0.9} & {\textbf{mIoU}} & {\textbf{RC} \textdownarrow}  \\
\midrule
\multirow{8}{*}{\makecell[c]{Biased \\ Models}} & \multirow{2}{*}{2D-TAN~\cite{zhang2020learning}} & high bias & 93.82 & 87.08 & 72.55 & 35.06 & 76.41 & \multirow{2}{*}{10.85} \\ 
& & low bias & 84.40\textsuperscript{\textbf{\fontsize{6pt}{6pt}\selectfont (-9.42)}} & 76.10\textsuperscript{\textbf{\fontsize{6pt}{6pt}\selectfont (-10.98)}} & 60.75\textsuperscript{\textbf{\fontsize{6pt}{6pt}\selectfont (-11.8)}} & 22.75\textsuperscript{\textbf{\fontsize{6pt}{6pt}\selectfont (-12.31)}} & 66.66\textsuperscript{\textbf{\fontsize{6pt}{6pt}\selectfont (-9.75)}} & \\
& \multirow{2}{*}{VSLNet~\cite{zhang2020span}} & high bias & 98.14 & 97.03 & 95.26 & 83.40 & 92.63 & \multirow{2}{*}{14.31}  \\
& & low bias & 85.59\textsuperscript{\textbf{\fontsize{6pt}{6pt}\selectfont (-12.55)}} & 83.22\textsuperscript{\textbf{\fontsize{6pt}{6pt}\selectfont (-13.81)}} & 79.60\textsuperscript{\textbf{\fontsize{6pt}{6pt}\selectfont (-15.66)}} & 67.34\textsuperscript{\textbf{\fontsize{6pt}{6pt}\selectfont (-16.06)}} & 79.16\textsuperscript{\textbf{\fontsize{6pt}{6pt}\selectfont (-13.47)}} & \\
& \multirow{2}{*}{LGI~\cite{mun2020local}} & high bias & 97.02 & 94.26 & 87.38 & 56.36 & 85.25 & \multirow{2}{*}{14.94}   \\
& & low bias & 89.70\textsuperscript{\textbf{\fontsize{6pt}{6pt}\selectfont (-7.32)}} & 82.98\textsuperscript{\textbf{\fontsize{6pt}{6pt}\selectfont (-11.28)}} & 68.74\textsuperscript{\textbf{\fontsize{6pt}{6pt}\selectfont (-18.64)}} & 31.49\textsuperscript{\textbf{\fontsize{6pt}{6pt}\selectfont (-24.87)}} & 72.67\textsuperscript{\textbf{\fontsize{6pt}{6pt}\selectfont (-12.58)}} & \\
& \multirow{2}{*}{QD-DETR~\cite{moon2023query}} &  high bias & 98.96 & 98.35 & 96.46 & 82.61 & 93.05 & \multirow{2}{*}{5.92} \\
& & low bias & 95.59\textsuperscript{\textbf{\fontsize{6pt}{6pt}\selectfont (-3.37)}} & 93.93\textsuperscript{\textbf{\fontsize{6pt}{6pt}\selectfont (-4.42)}} & 90.17\textsuperscript{\textbf{\fontsize{6pt}{6pt}\selectfont (-6.29)}} & 72.43\textsuperscript{\textbf{\fontsize{6pt}{6pt}\selectfont (-10.18)}} & 87.72\textsuperscript{\textbf{\fontsize{6pt}{6pt}\selectfont (-5.33)}} & \\
\midrule
\multirow{4}{*}{\makecell[c]{Debiased \\ Models}}& \multirow{2}{*}{DCM~\cite{yang2021deconfounded}} & high bias& 92.89 & 85.72 & 69.75 & 32.29 & 74.85 & \multirow{2}{*}{17.86} \\
& & low bias & 79.55\textsuperscript{\textbf{\fontsize{6pt}{6pt}\selectfont (-13.34)}} & 68.11\textsuperscript{\textbf{\fontsize{6pt}{6pt}\selectfont (-17.61)}} & 46.15\textsuperscript{\textbf{\fontsize{6pt}{6pt}\selectfont (-23.6)}} & 13.49\textsuperscript{\textbf{\fontsize{6pt}{6pt}\selectfont (-18.8)}} & 58.88\textsuperscript{\textbf{\fontsize{6pt}{6pt}\selectfont (-15.97)}} & \\
& \multirow{2}{*}{Shuffling~\cite{hao2022can}} &  high bias & 93.78 & 89.43 & 82.25 & 49.63 & 81.62 & \multirow{2}{*}{1.04}  \\
& & low bias & 93.26\textsuperscript{\textbf{\fontsize{6pt}{6pt}\selectfont (-0.52)}} &88.61\textsuperscript{\textbf{\fontsize{6pt}{6pt}\selectfont (-0.82)}} & 80.23\textsuperscript{\textbf{\fontsize{6pt}{6pt}\selectfont (-2.02)}}  & 49.04\textsuperscript{\textbf{\fontsize{6pt}{6pt}\selectfont (-0.59)}} & 80.36\textsuperscript{\textbf{\fontsize{6pt}{6pt}\selectfont (-1.26)}} &  \\
\bottomrule
\end{tabular}    
\end{center}
\vspace{-8mm}
\end{table*}

\subsection{Results and Analysis}

\noindent \textbf{Temporal Question Answering} The results of VidLLMs on SVLTA are shown in Table~\ref{tab:vidllms_comp}, indicating that none of the current VidLLMs can achieve satisfactory performance on our SVLTA benchmark, even some time-sensitive and close-sourced models. 
Specifically, the VTimeLLM only obtains the highest mIoU of $10.29$ among these time-aware VidLLMs and current strong close-sourced models like Gemini 1.5 Pro and GPT-4o just get the mIoU of $12.48$ and $18.90$, respectively. This means that current VidLLMs do not have strong temporal alignment capabilities. 
Additionally, we can observe that most general open-sourced VidLLMs often have poor temporal alignment ability such as Videochat2 and Video-LLaVA merely achieve the mIoU of $0.87$ and $2.59$ correspondingly, demonstrating that their training stage ignore the temporal understanding capability modeling. However, the Video-LLaMA2 has a mIoU of $12.33$, even higher than the time-aware VidLLMs, this is because of its temporal encoding design and high-resolution frame input. 
Further analysis of the visual domain gap, the number of frames, performance comparisons, and detailed question prompts is provided in \emph{Supplementary}.

\noindent \textbf{Distributional Shift Sensitiveness} 
The results in Table~\ref{tab:temporal_bias_sensitive} show the diagnosis of various specific temporal alignment methods in the distribution shift scenario. Notably, DCM, despite using causal inferencee~\cite{pearl2016causal,pearl2018book} to mitigate temporal bias effects, exhibits poorer robustness than biased methods (has the highest RC value of $17.86$). This suggests that the causal-based approach may have limitations in fine-grained shifts of temporal distribution, possibly due to the imperfect disentanglement of action content and position in videos. In contrast, Shuffling demonstrates better robustness (only has the lowest RC value of $1.04$), highlighting the effectiveness of using pseudo labels for video data augmentation to balance temporal distribution.
It generally shows weak result consistency regarding biased methods due to the lack of debiasing and inadvertently learning these biases. However, QD-DETR, a transformer-based model, outperforms other biased methods in robustness (has the lowest RC value of $5.92$ among the biased models), indicating superior generalization capabilities of transformer architectures. 

\begin{table}
  \footnotesize
  \centering
  \caption{The results of temporal alignment adaptation task.}
  \vspace{-6mm}
  \label{tab:syn2real}
  \begin{center}
  \setlength{\tabcolsep}{10pt}
    \begin{tabular}{@{}ccccc@{}}
    \toprule
    & \multicolumn{3}{c}{\textbf{R@1}} &  \\
    \cmidrule(lr){2-4} 
    \textbf{Method}  & IoU=0.3 & IoU=0.5 & IoU=0.7 & \textbf{mIoU}   \\
    \midrule
    2D-TAN~\cite{zhang2020learning} & 15.81 & 5.03 & 1.94 & 11.8  \\ 
    VSLNet~\cite{zhang2020span} & 28.33 & 8.52 & 3.87 & 19.66 \\
    LGI~\cite{mun2020local} & 33.96 & 12.52 & 3.30 & 22.24 \\
    QD-DETR~\cite{moon2023query} & 33.74 & 18.39 & 7.55 & 22.32  \\
    \bottomrule
    \end{tabular}
    \end{center}
    \vspace{-8mm}
\end{table}

\noindent \textbf{Temporal Alignment Adaptation} The results are illustrated in Table~\ref{tab:syn2real} and we can observe: 1) several frameworks of alignment models can transfer temporal knowledge (e.g., VSLNet and LGI can achieve the mIoU of $19.66$ and $22.24$ respectively). It means these models trained from scratch can transfer their temporal alignment ability to the new situations or domains, 2) transformer-based model has better transferability than other frameworks, it can achieve $10.52$ higher mIoU than 2D-TAN and $2.66$ than VSLNet, which demonstrates the advantages of transformer architectures in temporal alignment when adapting to new situations.

\section{Conclusion}
In this work, we first systematically analyze the temporal distributions for the vision-language temporal alignment problem from the decomposition aspect and introduce a new metric TJSD to examine three specific types of temporal bias related to process, entity, and composition. After that, we build a new large-scale and compositional benchmark SVLTA, using a proposed synthetic pipeline. 
Our approach involves activity commonsense, controllable activity manuscript, and constrained filtering to ensure it is diverse, compositional, and unbiased.  
The experiments reveal interesting insights for using this dataset in various diagnostic tasks, such as temporal question answering, distributional shift sensitiveness, and temporal alignment adaptation.


{
    \small
    \bibliographystyle{ieeenat_fullname}
    \bibliography{main}
}


\end{document}